\documentclass[10pt,twocolumn,letterpaper]{article}

\usepackage{cvpr}
\usepackage{times}
\usepackage{epsfig}
\usepackage{graphicx}
\usepackage{amsmath}
\usepackage{amssymb}
\usepackage{multirow}

\usepackage[breaklinks=true,bookmarks=false]{hyperref}

\cvprfinalcopy 

\def\cvprPaperID{****} 

\begin{document}

\title{Attributes-aided Part Detection and Refinement for Person Re-identification}


\author{Shuzhao Li,
 Huimin Yu,
 Wei Huang,
 Jing Zhang \\
 \centerline{Zhejiang University} \\
 {\tt\small \{leeshuz, yhm2005, huangwayne28, zj9301\}@zju.edu.cn}
}

\maketitle

\begin{abstract}
Person attributes are often exploited as mid-level human semantic information to help promote the performance of person re-identification task. In this paper, unlike most existing methods simply taking attribute learning as a classification problem, we perform it in a different way with the motivation that attributes are related to specific local regions, which refers to the perceptual ability of attributes. We utilize the process of attribute detection to generate corresponding attribute-part detectors, whose invariance to many influences like poses and camera views can be guaranteed. With detected local part regions, our model extracts local features to handle the body part misalignment problem, which is another major challenge for person re-identification. The local descriptors are further refined by fused attribute information to eliminate interferences caused by detection deviation. Extensive experiments on two popular benchmarks with attribute annotations demonstrate the effectiveness of our model and competitive performance compared with state-of-the-art algorithms.
\end{abstract}

\section{Introduction}

Person Re-identification (Person Re-ID) has become a very popular and challenging topic during recent years. Given a pedestrian image of interest called ``probe image", the Re-ID task aims to search in a large gallery image database for images of the same identity as the probe, which can also be treated as an image retrieval task. It is of great importance in both research fields and video surveillance applications.

Despite many years of researches on Re-ID task, it is still an issue full of challenges. Firstly, since the probe and gallery images are taken under non-overlapped cameras, the large variations of visual perspectives, illuminations and poses can be very confusing when making a judgment on whether two images contain the same identity. Secondly, since the human body regions are detected by existing object detection methods such as DPM~\cite{felzenszwalb2010object} or Faster-RCNN~\cite{ren2015faster}, the detected bounding boxes may be inaccurate, which together with the pose variations, to cause the problem of spatial misalignment between two images. Apart from these, the occlusion problem frequently occurring in realistic video surveillance scenes may cause the absence of important clues for identifying someone. Considering the challenges above, learning a view-invariant and robust feature expression is essential for an effective person Re-ID system.

\begin{figure}
\begin{center}
    \includegraphics[width=0.9\linewidth]{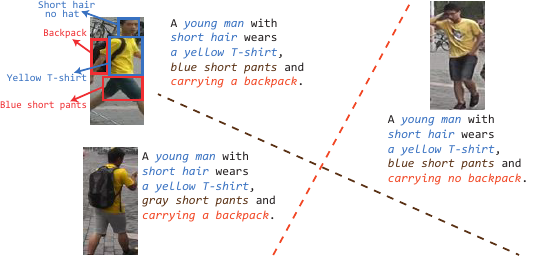}
\end{center}
    \caption{Attributes are consistent with human recognition mechanism for identifying people. The pedestrian can be described by a combination of various different kinds of attributes and some of these attributes can be discriminative for distinguishing between similar but different persons.}
    \label{first graph}
\end{figure}

With increasing popularity of deep neural networks, more and more researchers tend to employ deep learning based feature representations rather than hand-crafted features for its excellent generalization on unseen data. After pretraining on ImageNet, we can easily acquire relatively powerful global feature representation for target person Re-ID task by further fine-tuning on its own training sets. While simple and effective, these global appearance representations cannot settle the misalignment problem described above for their lack of semantic perception for human body parts. Recently, popular researches over human pose estimation and attention models~\cite{vaswani2017attention,wei2016convolutional} have inspired many ideas for locating human body parts to alleviate the problems of pose misalignment and background interference. However, these methods either rely heavily on existing pose detection models which may arise errors when turned to Re-ID task, or use attention model as a part detector which may be hard to train due to the shortage of prior semantic knowledge as supervision.

Attribute learning for Person Re-ID task has been studied in recent years and proven to be of great help when treated as one kind of mid-level semantic features for its invariance to many influences like pose, camera views and lighting conditions. Suppose that we need to identify the man in the upper left corner of Figure~\ref{first graph}, we usually form a description like ``\textit{a young man wearing yellow T-shirt and blue short pants, carrying a backpack}". The description is fully made up of attribute information which indicates the consistency of attribute detection with human cognitive mechanisms. While in most existing approaches, attribute information is often simply incorporated into global feature learning by designing corresponding attribute classifiers~\cite{lin2017improving,matsukawa2016person}, they reckon without two important clues. On one hand, most attributes are associated with local regions and different from the holistic image-level feature representation, joint learning of these two kinds of features may cause the heteroscedasticity (a mixture of different knowledge granularity and characteristics) learning problem~\cite{duin2004linear} analysed in~\cite{wang2018transferable}. On the other hand, the description of a stranger is often a combination of several kinds of attributes while some of them are insignificant for re-identification, thus the relationship mining and selection of different attributes are vital for a robust learning.

To deal with these issues, in this paper we propose a deep learning based work called Attributes-aided Part Detection and Refinement model (APDR) which incorporates the attribute learning process in a different way. We wish that the learning process of attributes should have the perception for local human body regions and can be used as a pose-invariant part detector due to its invariance to many influences like human poses and camera views. Being exploited as prior knowledge when localizing body parts, attribute information makes the part detectors easier to learn than those attention models. Compared with the approaches using existing pose estimation models, attribute detection is directly optimized for the Re-ID task to avoid model deviation phenomenon and can provide extra semantic information in the meantime. Furthermore, attribute learning can detect regions and objects like \textit{handbag} or \textit{hat} which may be distinctive for identifying while pose models cannot. Therefore, the attribute localizer can be considered as a combination of attention model, human part detection model and salient object detection model. After that, in order to make our model work more like human experts that considering the relationship among attributes when identifying, a simple attribute fusion module is adopted to combine different kinds of attribute information. Taking it as a guidance, we make refinement on local part features extracted by the localizers to filter out the redundant and irrelevant interference introduced by the learned masks, which results in a powerful refined local descriptor for re-identification. The learned local features, along with holistic image-level feature, can further improve the accuracy on person re-identification task. We evaluate our APDR model on two public datasets with attribute annotations to verify the effectiveness of our ideas and demonstrate that our model can achieve state-of-the-art performance compared with other person Re-ID models.

The main contributions of our work are summarized as follows: (1) We propose a novel deep model called Attributes-aided Part detection and Refinement network (APDR) to firstly utilize the attribute learning process as a part localizer, which handles the part misalignment problem. To our best knowledge, it is the first time that the perceptual ability of attribute learning is explicitly integrated into person Re-ID task. (2) We design a simple but effective attribute fusion network to simulate the human behaviors of identifying people through attributes. (3) The fused attribute information is exploited as a guidance to filter out useless information to refine part features for a better representation.

\section{Related Works}
\textbf{Person Re-ID.} Most popular person Re-ID algorithms can be categorized into two classes, feature representation learning and metric learning. For the first category, usually the human identity labels are exploited as the supervision for training a classifier for different identities and can be considered as a classification problem. During recent years, CNN-based feature representation learning has been dominating various research fields because of its excellent performance and is no exception in person Re-ID community. Xiao \etal~\cite{xiao2016learning} propose a joint learning strategy to train a single classifier for multiple domains at the same time, and then fine-tune to adapt to each single domain with a domain guided dropout policy.

For metric learning methods, the similarity between different samples is compared for person matching. The input of a deep neural network is often in the form of image pairs or triplets~\cite{ahmed2015improved,li2014deepreid,varior2016gated}. The model will pull the feature distance of the same identity and push the distance between two different identities during learning process. Varior \etal~\cite{varior2016gated} design a gating function inserted in each CNN layer to compare multi-scale similarities between the input image pair. Zheng \etal~\cite{zheng2017discriminatively} design two types of classifiers which combine feature learning and metric learning together. Though simple and effective, these models with holistic image-level features did not take the human part misalignment problem into consideration.

\begin{figure*}
\begin{center}
\includegraphics[width=0.95\linewidth]{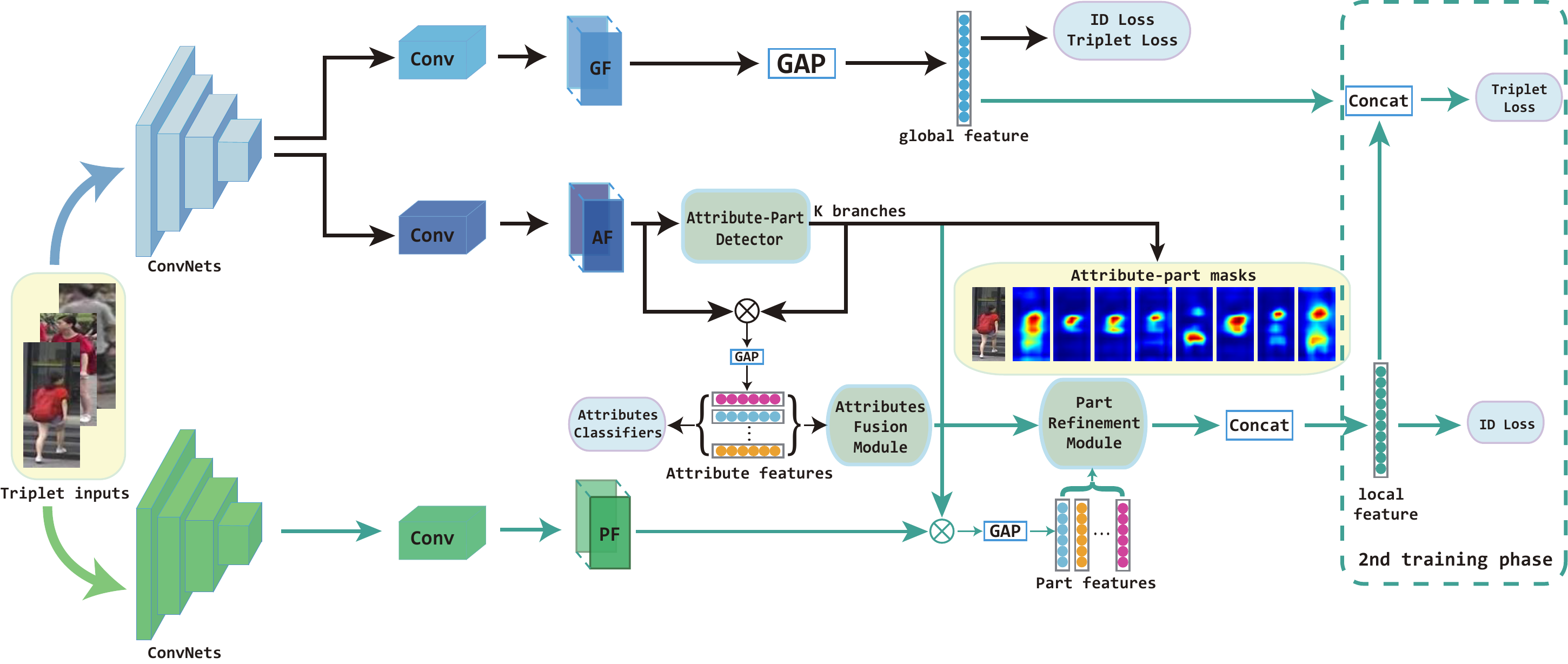}
\end{center}
\caption{The overall architecture of our proposed APDR model. The whole model consists of a two-steam network and is correlated by the part refinement module. For an input triplet, each image will firstly go through the upper stream to perform identity and attribute learning, several attribute-part detectors for corresponding attributes are learned to make full use of the perceptual ability of attributes. For the stream below, the model will utilize the learned detectors to extract part features, and guided by the fused attribute representation, the part feature can be further refined. Together with global feature, we acquire the final feature representation for our APDR model. (Best viewed in color.)}
\end{figure*}

\textbf{Body part-aligned representations.} To deal with the part misalignment problem in person Re-ID, more and more algorithms focus on extracting local human part feature for re-identification and can be classified into two categories. One is to employ existing human pose estimation algorithms to locate the body parts. Some researchers~\cite{wei2016convolutional,zhao2017spindle,zheng2017pose} adopt CPM~\cite{wei2016convolutional} to predict human body joints and generate body regions to extract part features, Zhao \etal~\cite{zhao2017spindle} then design a tree-structured feature fusion strategy to merge different part features to form the final feature. Zheng \etal~\cite{zheng2017pose} and Su \etal~\cite{su2017pose} rearrange human part patches to generate a new pose-aligned human image. These methods rely heavily on the detection accuracy of existing models trained for other tasks.

The other category is based on attention models to locate human parts or salient regions, which can be considered as an unsupervised manner. Liu \etal~\cite{liu2017end} exploit LSTMs as attention modules to locate different human attention parts, and Zhao \etal~\cite{zhao2017deeply} learn several human part maps supervised only by triplet loss for re-identification. These models are simple to construct but hard to train because the supervision for re-identification is too weak for part detectors to learn effective salient regions.

\textbf{Attribute learning.} Treated as one kind of middle and high level semantic feature, attributes have provided many valuable auxiliary information for person Re-ID. Su \etal~\cite{su2016deep} train an attribute learning model treating the deep attribute feature as the final representation for person matching which ignores the fact that different people may share similar attributes. Matsukawa and Suzuki~\cite{matsukawa2016person} propose new labels by combining different attribute labels to train extra classifiers in addition to single-attribute labels. Lin \textit{et al.}~\cite{lin2017improving} have provided the attribute annotations for two large-scale person Re-ID datasets Market-1501~\cite{zheng2015scalable} and DukeMTMC-reID~\cite{zheng2017unlabeled}. The algorithms above all treat the attribute learning as a simple feature extraction process, while ignoring the perceptual ability of attributes.

\section{Perceptual attribute detection}\label{perceptual attribute detection}
Person attribute learning has been studied a lot in recent years, and has been proven beneficial for the person Re-ID task. The human attributes can be grouped into two categories, one of which is corresponding to local parts of the human body or certain regions of an image such as \textit{T-shirt} or \textit{backpack}, and the other is high-level semantic attributes that cannot be assigned to specific region of human body or can be considered as associated with the whole human body like \textit{age} and \textit{gender}. Though containing global information, it is different from the holistic image-level feature for its independence of background. Briefly speaking, apart from the auxiliary semantic information brought by attributes, the procedure of attribute detection will concentrate on discriminative human body parts and salient objects contained in an input image.

Since the process of attribute detection and part localization can be done at the same time, the motivation of our work is to train several attribute-part detectors to fully utilize the perceptual ability of human attributes. Compared with previous approaches based on pose or attention models, our attribute-part detectors learned through the perception of human attributes are easier to train and are directly optimized for the person ReID task. Motivated by these, we propose a simple attribute detection network based on ResNet-50 model~\cite{he2016deep}. As analysed in \cite{wang2018transferable}, co-learning attribute and identity tasks can be beneficial for both tasks, while different from the architecture in \cite{wang2018transferable} or \cite{lin2017improving}, we separate the attribute and image-level feature learning into two branches after ``pool4" block to avoid the heteroscedasticity problem~\cite{wang2018transferable}. For image-level feature learning branch, we simply replace the last $7\times7$ pooling operation by global average pooling (GAP) to accommodate to different input resolutions. For attribute learning branch, we remove the last spatial down-sampling layer of the backbone network to increase the resolution of the final feature map for preserving more details, which is beneficial for the further attribute learning and localization. Let $x$ denote the input image, $\mathbf{G}$ and $\mathbf{A}$ represent the global image-level feature extraction branch and attribute learning branch, the corresponding feature maps are obtained by:
\begin{align}
\mathbf{GF} = {\mathbf{G}(x; {\theta}_{g})}, \qquad \mathbf{AF} = {\mathbf{A}(x; {\theta}_{a})}
\end{align}
\begin{align}
\mathbf{g} = {\mathcal{GAP}(\mathbf{GF})}
\end{align}
where ${\theta}_{g}$ and ${\theta}_{a}$ represent the parameters in backbone network, $\mathbf{g}$ is the global feature output by image-level feature learning branch. After acquiring the attribute feature map $\mathbf{AF}$, we use it to learn several attribute-part detectors to generate the attention mask for each attribute:
\begin{align}
\mathbf{M}_i = N_{attri\_detector_i}(\mathbf{AF})  \label{attention weights}
\end{align}
where $N_{attri\_detector_i}(*)$ is the attribute detector composed of a Convolution and Sigmoid operation to normalize attention scores for each location. With the generated masks, we further extract each attribute feature by performing weighted average pooling operation over all locations on the attribute feature map, whose weights are given by attention masks in Eq.~\ref{attention weights}.
\begin{align}
\mathbf{m}_i &= \frac{\sum_{(x, y)}\mathbf{AF}(x, y) * \mathbf{M}_i(x, y)}{H*W}
\end{align}
\begin{align}
\mathbf{a}_i &= \mathcal{BN}(\mathcal{FC}(\mathbf{m}_i))
\end{align}
where $\mathbf{m}_i$ denotes the pooled feature of the $i$th attribute, $\mathbf{AF}(x, y)$ is the $c$-dim feature vector of location $(x, y)$ on attribute feature map, $H$ and $W$ denote the size of the feature map. The averaged feature is further sent into a linear dimension-reduction layer along with a Batch Normalization layer to obtain the final attribute feature $\mathbf{a}_i$. Finally, they are sent to their corresponding classifiers, supervised by the annotated attribute labels with the cross-entropy loss.
\begin{gather}
\begin{split}
\mathbf{L}_{attri} = \sum\limits_{i=1}^N {p}_{i} * {\log} {q}_{i}
\end{split}
\end{gather}
where $N$ is the number of attributes, $q_i$ is the predicted probability for target class $t$ of the $i$th attribute and $p_i=1$ for its corresponding ground-truth class.

Considering that different attributes may target at the same or similar human body regions, we manually merge these attributes to share the same attribute attention mask while generating separate feature representations for them. Another motivation to merge the location-shared attributes is to increase the training samples for the attribute-part detector in that some attributes like \textit{wearing hat or not} may be hard to learn and detect because of the huge imbalance between positive and negative samples in common scenarios, but when learned together with \textit{long/short hair} attribute, it is easier for our model to obtain a head-region part detector. Hence, we design $K = 8$ attribute detectors for our model and are corresponding to \textit{age}, \textit{backpack}, \textit{bag}, \textit{handbag}, \textit{lower body}, \textit{upper body}, \textit{head}, \textit{gender} for Market-1501. More details are described in Section \ref{sec:ablation}

For a better optimization for re-identification task, we adopt both human identity supervision and triplet metric constraints on the image-level feature:
\begin{align}
\mathbf{L}_{id} = \frac{1}{L}\sum\limits_{i=1}^L p_{\mathbf{g}_i}*\log(q_{\mathbf{g}_i})
\end{align}
\begin{align}
\mathbf{L}_{tri} = \frac{1}{M}\sum\limits_{i=1}^M [{d_i^p} - {d_i^n} + m]_{+}
\end{align}
where $L$, $M$ denote the number of identities and triplets within a batch, $[*]_{+} = \max(*, 0)$ is the hinge loss, $d_i^p = \Vert{\mathbf{g}_i^a} - {\mathbf{g}_i^p}\Vert_2^2, d_i^n = \Vert{\mathbf{g}_i^a} - {\mathbf{g}_i^n}\Vert_2^2$ are the distances of positive and negative pairs, $m$ is the margin to separate them. The final loss for the perceptual attribute detection is composed of three terms:
\begin{align}
\mathbf{L} = {\mathbf{L}_{id}} + {\mathbf{L}_{tri}} + {\lambda}{\mathbf{L}_{attri}}
\end{align}
where the parameter ${\lambda}$ is determined by cross-validation to balance the attribute and identity learning to avoid over-fitting.

\section{Part Feature Refinement}  \label{sec:part feature refinement}

When people are re-identifying someone, they usually pick out the most discriminative attributes such as \textit{red coat} or \textit{white backpack}, and neglect some common attributes which are not helpful for identification. The relationship of attributes is also important since we always combine different kinds of attributes to perform recognition. Hence, we design a simple fusion network to build the relationship of various attributes. The module is composed of a dense layer to aggregate all attribute information into a single vector, the attributes can be merged and selected through trainable weights ${\theta_f}$.
\begin{align}
\mathbf{f}_{attri} = \mathcal{FC}(\mathbf{a}_1, \mathbf{a}_2, ..., \mathbf{a}_N; {\theta}_{f})
\end{align}

Besides this, as described in the section before, people with different identities may share similar attributes that can confuse the judgment of our model, so re-identifying person by directly comparing the attribute information may not be appropriate. Considering that despite sharing similar attributes, the corresponding local features of those parts should differ from each other due to their different identities. Based on this observation, we make use of the attribute-part detectors obtained in the previous section to extract discriminative human part features. For the reason that attention regions are learned through attribute labels whose supervision for localization is a little weak, these regions may overlap with each other and contain some irrelevant background information, so we take the fused attribute feature as a guidance to filter out insignificant components and refine the part features for a more robust representation. The part feature refinement process can be seen as follows:
\begin{align}
\mathbf{l}_i = \frac{\sum_{(x, y)}\mathbf{PF}(x, y) * \mathbf{M}_i(x, y)}{H * W}
\end{align}
\begin{align}
\mathbf{p}_i = \mathbf{l}_i * \sigma(W_{pi}\tanh(W_{li}\mathbf{l}_i + W_{hi}\mathbf{f}_{attri} + b_i))
\end{align}
where $\mathbf{PF} = {\mathbf{P}(x; {\theta}_{p})}$ denotes the part feature map,  $\mathbf{l}_i$ and $\mathbf{p}_i$ denote the part feature before and after refined, $W_{*}$ is the linear transformation matrix. The refinement elements are calculated by correlating part features with the fused attribute information.

In summary, the motivation of designing the attribute fusion module lies in two aspects. On one hand, by aggregating different kinds of attribute information to form a mid-level human semantic feature, we want to simulate the recognition process of human beings to form a compact attribute descriptor. On the other hand, the fused attribute feature can serve as a guidance for the part feature refinement process, promoting the performance of the part branch.

All the refined part features $\mathbf{p}_i$ are concatenated and then dimension-reduced to form the final local feature. Together with the holistic feature, the final feature representation is shown in Eq.~\ref{final feature}, it contains three types of information: refined human part feature, fused attribute information and holistic image-level feature. We do not concatenate the fused attribute feature into our final feature representation for two reasons: one is that it is designed for aggregating attribute information but not directly optimized for distinguishing between different people, especially for those with similar attributes, the other is that during performing part refinement, the fused attribute information has been integrated into the local representations, whose effectiveness will be reflected in the final feature expression.

The part feature refinement module is optimized during the second training phase, where the local feature is supervised by identity loss and the final feature by triplet loss. For the reason that person Re-ID is more a distance metric task than classification task, only triplet loss is applied to make the global and local features cooperate well in final representation.
\begin{align}
\mathbf{f}_p = \mathcal{FC}(\mathbf{p}_1, \mathbf{p}_2, ... , \mathbf{p}_K; \theta_p)
\end{align}
\begin{align}
\mathbf{f} = [\mathbf{f}_p, \mathbf{g}] \label{final feature}
\end{align}

\section{Implementation Details}
We implement our proposed algorithm based on PyTorch framework on a GTX Titan Xp GPU with 12GB memory. We adopt ResNet-50 pretrained over ImageNet as our backbone network. The feature dimensions of holistic and part feature are both set to 256, thus to form the 512-d final feature. The number of learned attribute masks is set to 8 for both two datasets.

We exploit a two-stage training scheme, In the first stage, we perform perceptual attribute detection to simultaneously obtain the global image-level feature, individual attribute features and perceptual attribute masks. With learned attribute masks and individual attribute features, in the second training stage, we mainly optimize the attribute fusion module and part refinement module to obtain the refined local features.

The whole network is optimized using stochastic gradient descent (SGD) with momentum on mini-batches, the initial learning rate for the first training stage is set to 0.01 and decreased by 0.2 every 50 epochs. In the second training phase, the learning rate setting of attribute fusion and part feature refinement module is the same as above, while for the perceptual attribute learning module which has been optimized in the first stage, we set a small learning rate of 0.0001 to keep the learned features and masks basically unchanged for a stable learning. The hyper-parameters $m$ and $\lambda$ in the loss function are set to 0.2 and 0.1, the weight decay and momentum are set to 0.0005 and 0.9 respectively.

\begin{figure*}
    \centering
    \includegraphics[width=0.95\linewidth]{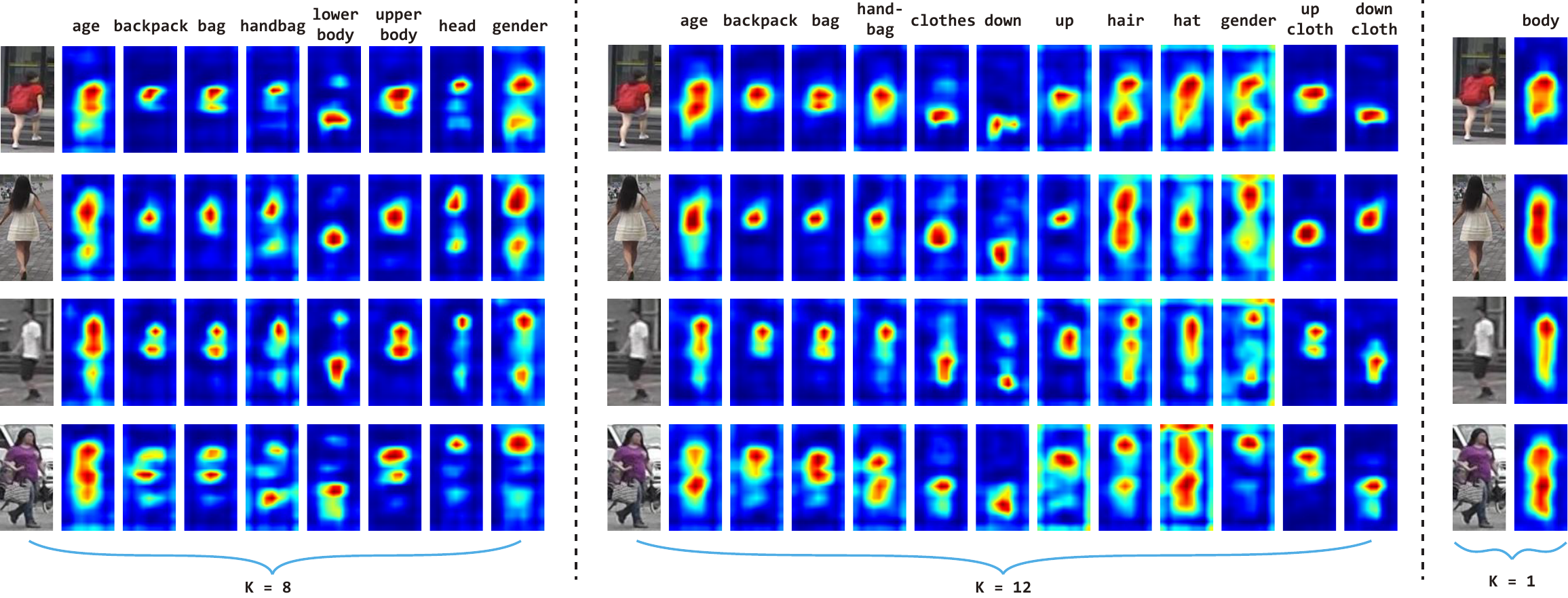}
    \caption{Examples of the attention masks learned in different settings. We adopt $K = 8$ in our APDR model. (Best viewed in color.)}
    \label{attribute maps}
\end{figure*}

\section{Experiments}
In this section, we report our experimental results on standard datasets and give a detailed ablation study over different modules of our APDR model. Extensive experiments are conducted on two large and challenging benchmarks: Market-1501~\cite{zheng2015scalable} and DukeMTMC-reID~\cite{ristani2016performance,zheng2017unlabeled}, which demonstrate that our approach is comparable to other state-of-the-art algorithms.

\subsection{Datasets and evaluation metric}
\textbf{Market-1501} is one of the largest and most challenging person ReID datasets lately. The original images are collected from 6 cameras in front of a supermarket in Tsinghua University and the pedestrian bounding boxes are cropped by Deformable Part Model (DPM)~\cite{felzenszwalb2010object}. The dataset contains 32668 annotated bounding boxes of 1501 identities, and 27 kinds of attribute labels for each identity. Among 1501 identities, 12936 images of 751 identities are partitioned for training and the rest 750 identities are left for testing.

\textbf{DukeMTMC-reID} is a subset of the DukeMTMC dataset~\cite{ristani2016performance} designed for person re-identification. It consists of 36411 human bounding boxes belonging to 1812 identities, among which 1404 identities appear in more than two cameras and 408 identities appear in only one camera treated as distractors. The dataset provides 23 kinds of attributes for each identity. The training set contains 16522 images of 702 identities and the rest 702 identities are assigned to the testing set.

Following the evaluation metrics widely used, we adopt both cumulative matching characteristics (CMC) and mean average precision (mAP) to evaluate our model under single query setting. The CMC score measures the accuracy of identifying the correct match at each rank. While for multiple ground truth matches in gallery, it can't tell how well all the ground-truth matching images in the gallery are ranked. To remedy this, we also report mAP scores of our model.

\subsection{Ablation study} \label{sec:ablation}

\begin{table*}
\centering
\caption{The validation performance with different numbers of attribute-part detectors and the comparisons with different baselines over two datasets.}
\label{ablation study}

\begin{tabular}{|c|c|c|c|c|c|c|c|c|}
  \hline
  \multirow{2}*{Models} & \multicolumn{4}{|c|}{Market-1501} & \multicolumn{4}{|c|}{DukeMTMC-reID} \\ \cline{2-9}
                                & rank-1 & rank-5 & rank-10 & mAP & rank-1 & rank-5 & rank-10 & mAP \\ \hline
  Baseline                      & 87.6 & 94.9 & 96.9 & 69.0 & 76.0 & 87.5 & 90.8 & 57.6 \\
  Baseline + Triplet Loss       & 88.6 & 95.8 & 97.3 & 71.9 & 79.7 & 89.4 & 92.1 & 62.1 \\
  1 mask for attributes         & 90.3 & 96.2 & 98.0 & 76.4 & 81.1 & 90.5 & 93.4 & 65.8 \\
  12/10 masks for attributes    & 90.5 & 96.5 & 97.7 & 76.7 & 80.8 & 90.1 & 93.2 & 65.9 \\
  Perceptual attribute learning & 91.3 & 96.5 & 97.9 & 77.0 & 82.0 & 91.2 & 93.8 & 66.4 \\  \hline
  Part branch                   & 89.5 & 96.4 & 97.8 & 73.7 & 80.5 & 90.5 & 93.2 & 64.6 \\
  Refined part branch           & 90.7 & 96.6 & 97.9 & 76.0 & 81.1 & 90.3 & 93.3 & 65.2 \\ \hline
  APDR (Ours)                   & 93.1 & 97.2 & 98.2 & 80.1 & 84.3 & 92.4 & 94.7 & 69.7 \\
  \hline
\end{tabular}
\end{table*}
\vspace{1ex}
\noindent\textbf{The number of attribute-part detectors.} We empirically study how the number of attribute-part detectors affects our model's performance. The number of annotated attribute labels in Market-1501 is 27, among which 8 and 9 attributes are respectively corresponding to colors of upper-body and lower-body clothing, so we first merge these to two multi-class attributes and form 12 kinds of attributes. Furthermore, as described in previous section, we merge attributes targeted at the same or similar regions to share one mask. We do not share the same mask between \textit{gender} and \textit{age} because although they can be considered as related to the whole body, they contain different high-level semantic meanings and may result in emphasizing on different regions. Finally, we form 8 different masks for the 12 kinds of attributes.

Based on above analysis, we conduct the experiment over Market-1501 dataset with different numbers of detectors $K = 1, 8, 12$. When $K = 1$, we simply generate one mask for all attributes and the mask can be seen as a detector for the whole body. The results are shown in Table~\ref{ablation study} and the learned masks are shown in Figure~\ref{attribute maps}. We find that the learned masks are able to concentrate on the whole human body regions to eliminate the influence of irrelevant background, while still miss some vital regions like heads and handbags. When $K = 12$, we can observe that some of the detected regions are similar to each other thus are redundant, and some learned masks cannot target at corresponding regions precisely due to the imbalanced training samples, like the one relevant to \textit{hat}.

When $K = 8$, which is denoted as our ``perceptual attribute learning", the learned 8 masks can concentrate on different salient regions and are highly consistent with their target attributes. Obviously, merging the \textit{hair} and \textit{hat} attribute to one \textit{head} mask yields more satisfying localization results than the $K = 12$ setting. We also find that the \textit{age} and \textit{gender} masks which correspond to high-level semantic attributes, though related to the whole body, have their own focuses. For the \textit{gender} mask, it prefers the regions of heads and lower body parts, we analyze the reason is that we usually judge the gender of someone firstly by observing his facial feature or hair length, the lower-body clothing can also be more discriminative for judging than upper-body clothing, such as `dress' for female and `pants' for male, compared to `T-shirt' for both genders. While the \textit{age} mask focuses mainly on the upper body region, which is probably because that human often estimate the age of someone depending on the style of upper clothing. In summary, with the appropriate choice for the number of attribute-part detectors, the performance is the best among the three settings and the learned masks are satisfactory.

For DukeMTMC-reID's 23 kinds of attributes, we also merge the 8 \textit{colors of upper-body clothing} and 7 \textit{colors of lower-body clothing} to two multi-class attributes, hence form 10 different kinds of attributes and design 8 detectors for them. The 8 detectors are corresponding to \textit{backpack}, \textit{bag}, \textit{handbag}, \textit{feet}, \textit{gender}, \textit{head}, \textit{upper body} and \textit{lower body}. Comparison results are shown in Table~\ref{ablation study}.

\vspace{1ex}
\noindent\textbf{Baseline comparison.} The comparisons of different baselines are listed in the first block of Table~\ref{ablation study}. Compared with the simple baseline which only adopts identity labels as supervision, our attribute learning baseline outperforms it by a large margin, which verifies the effectiveness of learning with attribute information. We add triplet loss to our baseline in that person re-identification is actually a metric learning problem. Taking the perceptual attribute learning as our strong baseline, the best performance is obtained, apart from which the learned attribute features and attention masks can be utilized for latter part feature extraction and refinement.

We also conduct a baseline experiment on sharing the same feature representation for both attribute and identity learning, but the model is hard to optimize and results in unstable accuracy, so we do not list it in Table \ref{ablation study}. This phenomenon verifies the existence of heteroscedasticity learning problem and the effectiveness of our baseline structure design which separates the two learning tasks. However, considering the correlation between attribute and identity information, the two branches share the network weights in shallow layers and moreover, the learned robust attribute information is later incorporated into the final feature representation. In this way, we make full use of the semantic attribute information and at the same time, the heteroscedasticity learning problem can be avoided.

\vspace{1ex}
\noindent\textbf{Effect of attribute fusion and part feature refinement.} We empirically analyze the effectiveness of the attribute fusion and part refinement module in our model. The motivation of designing the attribute fusion module has been described in Section \ref{sec:part feature refinement}, whose effectiveness will be reflected in the refined part features.

To further validate the effectiveness of refinement for part features, we firstly concatenate all initial part features to form the local descriptor and evaluate on test sets, whose results are denoted as ``part branch". From the results we find that using part features alone can outperform the ``baseline with triplet loss", which demonstrates the effectiveness of our learned attribute attention masks for discriminative local feature extraction. After refinement, the accuracy of the part branch is further improved. Finally, by concatenating the refined local feature and global feature, we get the results of our APDR model. It promotes the accuracy by a large margin compared with the attribute learning baseline, which is $1.8\%$ gain for rank-1 and $3.1\%$ for mAP on Market-1501,  $2.3\%$ for rank-1 and $3.3\%$ for mAP on DukeMTMC-reID.

\subsection{Comparison with the state-of-the-arts}
In this section, we present the results of comparison with several state-of-the-art algorithms. Since our proposed approach involves both attribute learning and part detection, we compare our model with both two types of algorithms.

\vspace{1ex}
\noindent\textbf{Results on Market-1501.} On Market-1501, we compare our proposed algorithm with many state-of-the-art approaches, including manual feature designing algorithms: LOMO+XQDA~\cite{liao2015person} and BoW~\cite{zheng2015scalable}, metric learning based algorithms: KISSME~\cite{koestinger2012large}, WARCA~\cite{jose2016scalable} and SCSP~\cite{cheng2016person}, attribute learning algorithms: Attribute-Person Recognition network (APR)~\cite{lin2017improving}, Attribute-Complementary Re-id Net (ACRN)~\cite{schumann2017person}, algorithms based on part detection: Part Aligned Deep Features (PADF)~\cite{zhao2017deeply}, Harmonious Attention Network (HA-CNN) \cite{li2018harmonious}, Refined Part Pooling (RPP)~\cite{sun2017beyond}, Attention-Aware Compositional Network (AACN) \cite{xu2018attention}, Part-Aligned Bilinear Representations (PABR)~\cite{suh2018part}.

\begin{table}
  \caption{comparisons with state-of-the-arts on Market-1501.}
  \label{SOTA-market}
\begin{tabular}{|c|c|c|c|c|}
  \hline
  \multirow{2}*{Methods} & \multicolumn{4}{|c|}{Market-1501}   \\ \cline{2-5}
                          & rank-1 & rank-5 & rank-10 & mAP \\ \hline
  LOMO+XQDA               & 43.8 &   -  &   -  &  - \\
  BoW                     & 35.8 & 52.4 & 60.3 & 14.8 \\
  KISSME                  & 44.4 & 63.9 & 72.2 & 20.8 \\
  WARCA                   & 45.2 & 68.2 & 76   & - \\
  SCSP                    & 51.9 & 72.0 & 79.0 & 26.4 \\ \hline
  ACRN                    & 83.6 & 92.6 & 95.3 & 62.6 \\
  APR                     & 84.3 & 93.2 & 95.2 & 64.7 \\    \hline
  PADF                    & 81.0 & 92.0 & 94.7 & 63.4 \\
  AACN                    & 85.9 & -    & -    & 66.87 \\
  HA-CNN                  & 91.2 & -    & -    & 75.7 \\
  PABR                    & 91.7 & 96.9 & 98.1 & 79.6 \\
  RPP                     & 93.8 & \textbf{97.5} & \textbf{98.5} & 81.6 \\ \hline
  APDR (Ours)             & 93.1 & 97.2 & 98.2 & 80.1  \\
  +re-ranking             & \textbf{94.4} & 97.0 & 97.9 & \textbf{90.2} \\
  \hline
\end{tabular}
\end{table}

The detailed comparison results are listed in Table \ref{SOTA-market}. We can observe from the table that our approach performs better than most state-of-the-art algorithms except a little lower than ``RPP" and achieves the best accuracy in terms of both rank-1 and mAP after applying re-ranking process~\cite{zhong2017re}, which is often adopted as a post-process algorithm for re-identification to further boost accuracy. Though simple to construct, the ``RPP" model cannot handle well the large pose variations which occur more frequently in DukeMTMC-reID thus results in an inferior performance, and the partitioned parts demonstrate less semantic meaning for specific human body parts than our model. Our model outperforms ``PABR", which also adopts a two-stream network architecture, by $1.4\%$ on rank-1, $0.5\%$ on mAP before re-ranking. It is worth noting that our algorithm performs much better than either attribute learning methods or most human part based algorithms. For the former, we fully exploit the perception ability of human attributes instead of simply classifying them correctly and for the later, the part detection based on attribute information is more reliable for Person Re-ID task than those based on existing pose models or attention models.

\vspace{1ex}
\noindent\textbf{Results on DukeMTMC-reID.} We also evaluate our algorithm on another large benchmark with attribute annotations. Compared with Market-1501, person images from DukeMTMC-reID have more variations in resolution and background due to more complex scene layout, resulting in a more challenging task. We compare our approach with APR~\cite{lin2017improving}, ACRN~\cite{schumann2017person}, HA-CNN \cite{li2018harmonious}, AACN \cite{xu2018attention}, RPP~\cite{sun2017beyond}, PABR~\cite{suh2018part} and Table \ref{SOTA-duke} reports the results. From the results we can find that our model performs equally well on the DukeMTMC-reID benchmark and achieve the state-of-the-art performance. Although the `dilation' structure is adopted to further improve the accuracy of the original ``PABR" model, our algorithm performs almost the same as the final ``PABR" model on rank-1. And on the whole, our algorithm achieves a better performance. Similar as on Market-1501, the accuracy can be further improved by a large margin after the re-ranking process.

\noindent
\begin{table}
\caption{comparisons with state-of-the-arts on DukeMTMC-reID.}
\label{SOTA-duke}
\begin{tabular}{|c|c|c|c|c|}
  \hline
  \multirow{2}*{Methods} & \multicolumn{4}{|c|}{DukeMTMC-reID}   \\ \cline{2-5}
                          & rank-1 & rank-5 & rank-10 & mAP \\ \hline
  APR                     & 70.7 &   -  &   -  & 51.9 \\
  ACRN                    & 72.6 & 84.8 & 88.9 & 52.0 \\ \hline
  AACN                    & 76.9 &   -  &   -  & 59.3 \\
  HA-CNN                  & 80.5 &   -  &   -  & 63.8 \\
  RPP                     & 83.3 &   -  &   -  & 69.2 \\
  PABR                    & 84.4 & 92.2 & 93.8 & 69.3 \\    \hline
  APDR (Ours)             & 84.3 & 92.4 & 94.7 & 69.7 \\
  +re-ranking             & \textbf{87.3} & \textbf{93.0} & \textbf{95.2} & \textbf{83.2} \\
  \hline
\end{tabular}
\end{table}

\section{Conclusion}
In this paper, we proposed a novel Attributes-aided Part Detection and Refinement model aiming to utilize the perceptual ability of attribute learning and solve the part misalignment problem in person Re-ID at the same time. We demonstrated that, the attribute information is associated with discriminative body parts and salient regions, thus can be exploited to generate attribute-part detectors. Besides, in order to simulate the human cognitive mechanism of dealing with multiple attributes, we merge attributes through a simple module for a more compact attribute representation. Taking the learned attribute-part detectors as part localizers, we extract and further refine the local part features guided by fused attribute information to eliminate the noises introduced by detection deviation. The experiments on two large popular benchmarks verified the effectiveness of our model. In future work, we will dig more into understanding human recognition mechanism for re-identifying people, including but not limited to attribute information.

{\small
\bibliographystyle{ieee}
\bibliography{reference}
}

\end{document}